\documentclass[preprint,12pt]{elsarticle}

\usepackage{graphicx}
\usepackage{amsmath,amssymb}
\usepackage{booktabs}
\usepackage{multirow}
\usepackage{algorithm}
\usepackage{algpseudocode}
\usepackage{subcaption}
\usepackage{textcomp}
\usepackage{url}
\usepackage{lineno}
\usepackage[hidelinks]{hyperref}
\usepackage{subcaption}
\usepackage{float} 
\graphicspath{{figures/}}
\newcommand{\figorplaceholder}[2]{%
  \IfFileExists{figures/#2}{\includegraphics[#1]{#2}}{%
    \fbox{\parbox[c][4cm][c]{0.8\linewidth}{\centering\ttfamily missing: \detokenize{figures/#2}}}}}
\newcommand{\botrule}{\bottomrule}

\journal{Computer Vision and Image Understanding}

\begin{document}

\begin{frontmatter}

\title{Improving OCR Accuracy Through Multi-Classifier Verification of Visually Similar Characters}

\author[a]{Nuzhat Khan}
\ead{khan.nuzhat@utm.my}

\author[a]{Ab Al-Hadi Ab Rahman\corref{cor1}}
\ead{hadi@utm.my}

\author[b]{Mohamad Anuar Kamaruddin\corref{cor2}}
\ead{anuarkamaruddin@usm.my}

\author[c]{Muhammad Paend Bakht}
\ead{paend.bakht@mu.ie}

\author[a]{Sana Fatima}
\ead{sana@graduate.utm.my}

\author[a]{Farhad Khodadadi}
\ead{khodadadisharifabad@graduate.utm.my}

\author[a]{Ibrahim Yousef Alshareef}
\ead{yousefm@graduate.utm.my}

\cortext[cor1]{Corresponding author.}
\cortext[cor2]{Corresponding author.}

\affiliation[a]{%
  organization={Faculty of Electrical Engineering, Universiti Teknologi Malaysia},
  city={Johor Bahru},
  postcode={81310},
  country={Malaysia}%
}

\affiliation[b]{%
  organization={School of Industrial Technology, Universiti Sains Malaysia},
  city={Pinang},
  postcode={11800},
  country={Malaysia}%
}

\affiliation[c]{%
  organization={School of Business, Maynooth University},
  city={Maynooth},
  postcode={W23 F2H6},
  country={Ireland}%
}

\begin{abstract}
Distinguishing visually similar characters is a persistent challenge for optical character recognition (OCR) systems, causing errors in critical applications of text digitization. While preprocessing techniques improve image quality and word-level post-processing corrects spelling errors, neither addresses character-level visual ambiguity, an intrinsic property of certain character pairs that limits even state-of-the-art OCR models. We present the Character Verification Model (CVM), a character-level post-processing framework that verifies and corrects ambiguous characters in OCR outputs using specialized binary classifiers trained on visually confusable pairs such as `O'/`0', `Z'/`2', and `S'/`5'. CVM operates independently of dictionary validation, making it suitable for non-lexical data including serial numbers, alphanumeric codes, and license plates. Experiments on three OCR engines (Tesseract, EasyOCR, TrOCR) and ten fonts with low to high complexity demonstrate accuracy improvements of 3.95--14.81 percentage points, with an 87.3\% correction rate for visually similar character errors and a false-correction rate of only 2.3\%. The modular architecture integrates with existing OCR pipelines without modification to underlying recognition models.
\end{abstract}

\begin{highlights}
\item Character-level verification corrects visually similar character errors in OCR
\item One binary Random Forest classifier separates each confusable character pair
\item Dictionary-free design suits serial numbers, alphanumeric codes and plates
\item Plugs into Tesseract, EasyOCR and TrOCR without retraining the engines
\item Accuracy gains of 3.95 to 14.81 points across ten fonts and three engines
\end{highlights}


\end{frontmatter}


\section{Introduction}\label{sec1}

Optical character recognition (OCR) has become foundational technology for digitizing printed and handwritten text~\cite{park2025ocr}. Modern OCR systems achieve high accuracy through three fundamental components: image preprocessing to enhance input quality~\cite{hakasin2025impact}, classification models for character recognition~\cite{francis2025comparison}, and post-processing methods to correct recognition errors~\cite{nguyen2021correction}.

\begin{figure}[H]
  \centering
  \includegraphics[width=\linewidth]{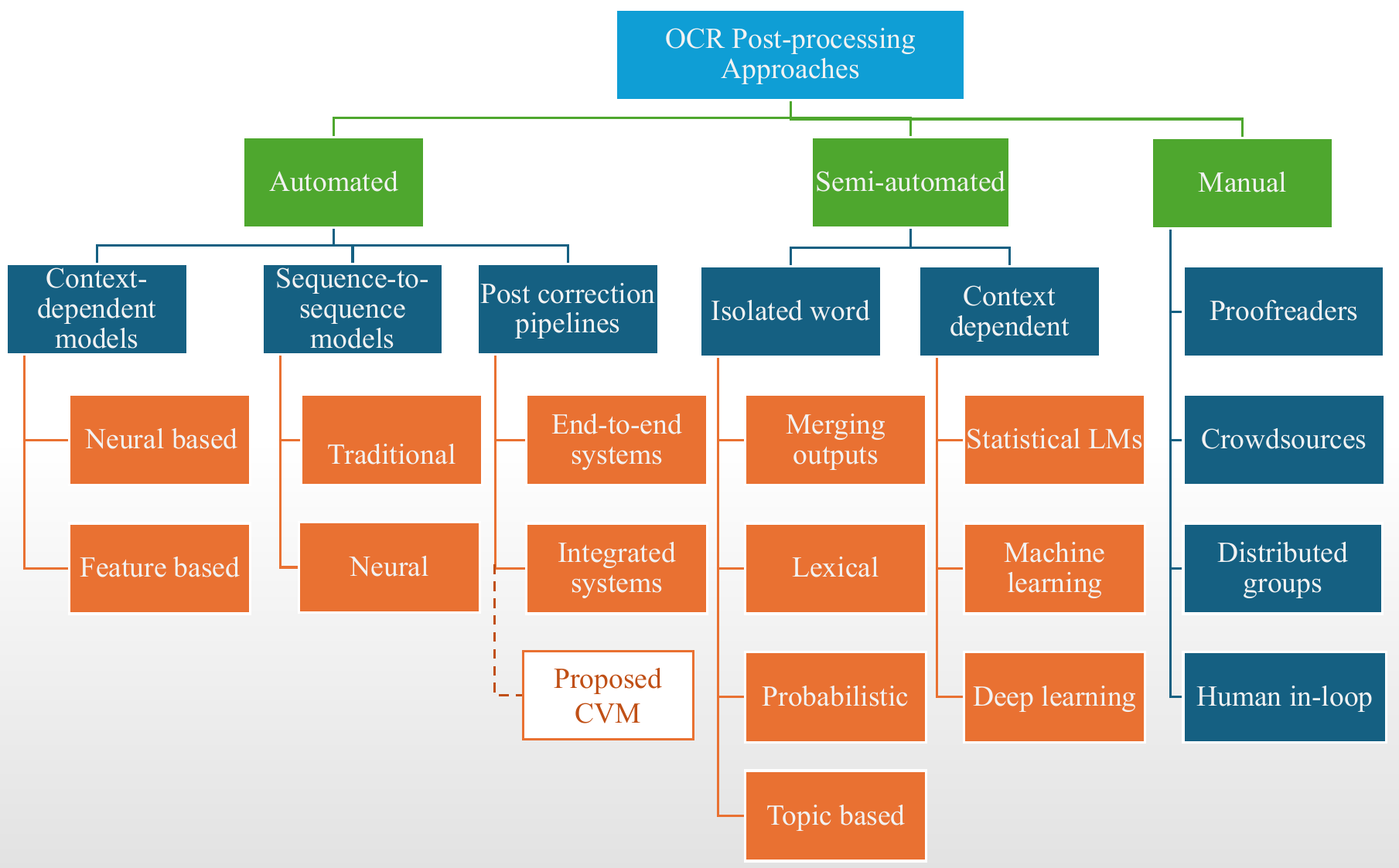}
  \caption{Taxonomy of post-processing methods for OCR error correction.}
  \label{fig:taxonomy}
    \end{figure}

Preprocessing techniques such as denoising, skew correction, binarization, and contrast enhancement mitigate noise and inconsistencies in input images~\cite{memon2020handwritten}. Classification models have evolved from traditional machine learning approaches to advanced deep learning architectures, hybrid models, and transformer-based systems incorporating multi-head attention~\cite{khan2025slr,zheng2024cdistnet}. Despite these advances, recognition errors remain unavoidable, particularly when characters share strong visual similarities~\cite{imam2022ocr}. Post-processing has therefore become an integral component of modern OCR pipelines~\cite{drobac2020optical}, evolving from manual correction and rule-based procedures to semi-automated approaches including dictionary-based correction, language-model-driven methods, and rule-based error detection (Fig.~\ref{fig:taxonomy}). More recently, fully automated post-processing models employing transformers and large language models have further improved error correction in digitized text~\cite{nguyen2021survey}. However, a fundamental limitation persists: visually similar characters (VSCs) remain difficult to distinguish even for state-of-the-art deep learning models~\cite{naseer2024investigating,oliveira2023evaluating,long2021scene}. Character pairs such as `O' and `0', `Z' and `2', or `S' and `5' share structural features that confound general-purpose classifiers, causing persistent misrecognition errors. This problem is particularly critical in safety- and business-critical applications, including autonomous navigation systems that rely on accurate street sign reading, license plate recognition for law enforcement, and inventory systems requiring precise serial number identification~\cite{meyers2023safety}.

Current post-processing methods are predominantly word-based and depend on lexical validity. Dictionary-based approaches perform well for natural language text but fundamentally fail when processing non-dictionary strings such as license plate identifiers, banking records, insurance claims, product serial numbers, alphanumeric codes, healthcare documents, and structured mixed-format data~\cite{gudi2025enhancing,paixao2018exploring,bansal2020necessary}. These strings do not conform to linguistic patterns and therefore fall outside the correction scope of conventional post-processing techniques. Similarly, standard OCR models trained to classify all characters simultaneously lack the fine-grained discriminative ability needed to separate confusable character pairs~\cite{gharib2022properness}.

Although word-level post-OCR correction has advanced from BERT-based error detection~\cite{nguyen2020neural} to hybrid TrOCR--language model systems~\cite{chen2024trocr} and recent large language model approaches~\cite{principe2024post}, these techniques operate at the word or sequence level and depend on linguistic context~\cite{nguyen2021survey}. At the same time, several OCR engines have achieved strong recognition performance through modern deep-learning pipelines. Tesseract~4.0 employs bidirectional LSTMs with CTC decoding and reports 97--98\% character accuracy on clean documents~\cite{smith2007overview,breuel2013high}. EasyOCR integrates CRAFT detection with a CRNN backbone (ResNet + BiLSTM + CTC), reaching competitive accuracy with real-time inference~\cite{baek2019character,shi2017end,easyocr2020}. Transformer-based OCR models such as TrOCR adopt a Vision Transformer encoder and a RoBERTa decoder, achieving a 2.89\% character error rate on IAM handwriting and high accuracy on receipt datasets~\cite{li2023trocr}. Recent developments, including Donut~\cite{kim2022ocr}, PaLI~\cite{chen2023pali}, and Nougat~\cite{blecher2023nougat}, further demonstrate the rapid progress of end-to-end transformer OCR systems.

Even with these advancements, visually similar character (VSC) confusion remains a persistent and underexplored challenge. Studies have shown that confusions involving characters such as \texttt{O/0}, \texttt{B/8}, \texttt{1/I/l}, and \texttt{2/Z/S} still occur in identity document processing~\cite{gunawan2021indonesian,alotaibi2025real}, license plate recognition~\cite{tagmaster2024license,kim2023super,nguyen2025deep,suharto2025comparative}, and answer sheet evaluation~\cite{kumar2024optical,vargoorani2024license}. These errors are critical in settings where text is non-lexical, domain-specific, or lacks contextual cues, making dictionary- or language-model-based correction unreliable. Despite the documented impact of VSC errors, no prior study has introduced a targeted character-level verification mechanism designed to detect and correct such confusions before the final OCR output. This gap motivates the need for a dedicated post-processing strategy that operates independently of linguistic context and directly addresses the visual similarity between characters.

\subsection{Related work}\label{subsec:related}

Post-OCR correction has progressed from rule-based methods to neural approaches over the past decade. Nguyen et al.~\cite{nguyen2021survey} categorized these approaches into isolated-word correction, context-dependent models, and sequence-to-sequence architectures, noting that modern methods fundamentally depend on lexical and contextual information. Traditional techniques employ edit distance algorithms and weighted finite-state transducers to map erroneous strings to dictionary entries~\cite{llobet2010ocr,reynaert2008non}, while recent transformer-based approaches have achieved state-of-the-art results. Nguyen et al.~\cite{nguyen2020neural} demonstrated that BERT-based models could detect and correct OCR errors by exploiting bidirectional context, winning the ICDAR 2019 competition. Large language models have also shown promise; Boros et al.~\cite{boros2024post} and Thomas et al.~\cite{thomas2024leveraging} explored GPT-based correction for historical documents, reducing character error rates by up to 38\%. However, these methods share a critical limitation: they require linguistic context and cannot validate non-dictionary strings such as serial numbers, passports, license plates, or alphanumeric codes.

The problem of visually similar character (VSC) confusion persists even in high-accuracy OCR systems. In license plate recognition, studies consistently report confusion between character pairs such as `B'/`8', `O'/`0', `I'/`1', and `Z'/`2'~\cite{laroca2018robust,silva2020real,goncalves2024license}. Gon\c{c}alves et al.~\cite{goncalves2024license} analyzed font characteristics contributing to these errors, identifying that similar apexes, crossbar positioning, and bowl shapes increase misrecognition likelihood. Notably, some jurisdictions have redesigned license plate fonts specifically to reduce such confusion~\cite{fe_schrift1994}, but OCR is still unable to recognize these character pairs. In document processing, similar VSC-related failures have been documented~\cite{hamdi2022depth,rigaud2019icdar}, with surveys identifying pairs like (O,0), (I,1), (S,5), (B,8), (Z,2) among the most problematic~\cite{memon2021automatic}. Despite widespread recognition of this problem, prior work has not addressed character-level visual similarity through targeted verification. Existing approaches either operate at the word level, requiring dictionary validation, attempt to improve the primary recognition model, or apply super-resolution techniques that do not resolve intrinsic visual ambiguity between character pairs~\cite{nascimento2023super}. This gap motivates our character verification approach, which operates independently of linguistic context and directly targets the visual ambiguity inherent in specific character pairs.

To address this gap, we propose the Character Verification Model (CVM), a character-level post-processing framework that operates as a secondary verification layer. Rather than validating entire strings or retraining the primary OCR engine, CVM employs specialized binary classifiers trained exclusively on confusable character pairs. By focusing computational effort precisely where errors are most likely, on VSC pairs that account for a substantial portion of OCR failures, CVM provides a lightweight yet highly effective correction mechanism.

The approach is grounded in the principle that verification is inherently easier than recognition. When the decision space is reduced from multi-class recognition (classifying among all possible characters) to binary comparison (distinguishing between two known confusable characters), specialized discriminators achieve significantly higher accuracy than general-purpose OCR models~\cite{delmoral2022multiclass}. With this design, CVM can correct errors that persist even after extensive preprocessing and classification by state-of-the-art OCR systems.

The main contributions of this work are as follows:
\begin{itemize}
\item We formulate character verification as a specialized binary classification problem and demonstrate that focused discriminators outperform general-purpose OCR models on visually similar character pairs.
\item We introduce a vocabulary-independent, character-level post-processing architecture suitable for correcting non-dictionary strings in safety-critical and business-critical applications.
\item We provide extensive empirical evidence across three OCR engines (Tesseract, EasyOCR, TrOCR) and ten fonts, demonstrating that CVM consistently reduces VSC-induced errors with accuracy improvements of 3.95--14.81 percentage points, an 87.3\% correction rate for VSC errors, and a false-correction rate of only 2.3\%.
\end{itemize}

The remainder of this paper is organized as follows: Section~\ref{sec:methods} presents the methodology, including problem formulation, dataset construction, CVM architecture, and training strategy; Section~\ref{sec:results} reports experimental results and discusses performance across different OCR engines and fonts; Section~\ref{sec:conclusion} concludes with key findings, limitations, and future research directions.

\section{Materials and methods}\label{sec:methods}

This section presents the CVM, beginning with problem formulation, followed by dataset construction, system architecture, training methodology, and the inference pipeline.

\subsection{Problem formulation}\label{subsec:formulation}

Let an OCR system generate an output string such as:
\begin{equation}
S = c_1 c_2 \ldots c_n
\label{eq:string}
\end{equation}
where $c_i$ denotes the predicted character at position $i$, and let the set of visually similar character (VSC) pairs be defined as:
\begin{equation}
\mathcal{P} = \{(a_j, b_j)\}_{j=1}^{m}
\label{eq:pairs}
\end{equation}
where each pair $(a_j, b_j)$ contains characters that frequently cause OCR confusion, such as alphanumeric pairs (O/0, Z/2, S/5, B/8, I/l, I/1), letter pairs (U/V, C/G, o/a, o/e, u/v, n/r, h/b, g/q), and symbol--character pairs (I/!, t/+). The complete set of confusable pairs is provided in the supplementary material.

The objective of CVM is to verify each character $c_i \in S$ that belongs to any confusable pair in $\mathcal{P}$ and determine whether it should be replaced with its visually similar counterpart, using a pair-specific classifier:
\begin{equation}
C_j : \mathbb{R}^{d} \rightarrow \{a_j, b_j\}
\label{eq:classifier}
\end{equation}
Formally, for each $c_i$ such that $\exists (a_j, b_j) \in \mathcal{P}$ where $c_i \in \{a_j, b_j\}$, a specialized binary classifier is applied to re-classify the character image patch $x_i$ extracted from the original input image.

\subsection{CVM training dataset construction and preprocessing}\label{subsec:dataset}

Character patches were derived from the publicly available 94-class ASCII OCR dataset\footnote{\url{https://github.com/VeCAD-P04-UTM/94-Class-ASCII-Image-OCR-Data}}, which contains segmented and normalized images of alphanumeric and special characters across multiple fonts. The dataset comprises character images in 50 different fonts with 3--5 samples per font, totaling over 3,000 samples across all character classes.

\begin{figure}[H]
  \centering
  \includegraphics[width=\linewidth]{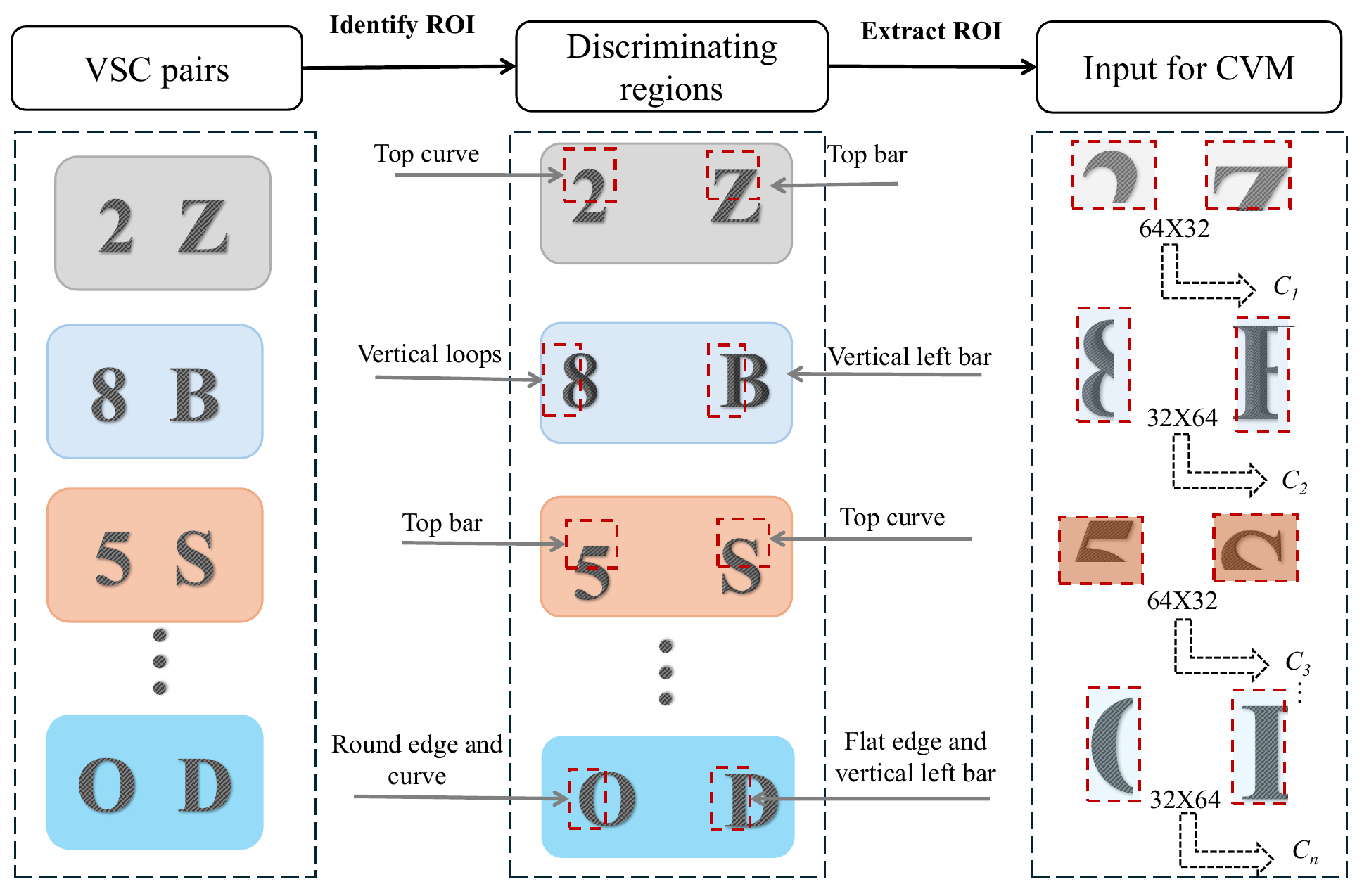}
  \caption{Character patch extraction process for binary classifier training.}
  \label{fig:roi}
\end{figure}

For each VSC pair in $\mathcal{P}$, each character image was divided into two discriminative regions of interest (ROIs), yielding approximately 300 patches per character and 600 patches per pair. The extraction process, illustrated in Fig.~\ref{fig:roi}, focused on identifying and isolating these ROIs within each character. They capture unique structural features such as curvature, stroke thickness, terminal shapes, and angles that distinguish confusable characters. For instance, distinguishing `0' from `O' requires analyzing edge sharpness and aspect ratio, while separating `S' from `5' relies on detecting the horizontal bar present in `5'.

\subsubsection{Character patch standardization}\label{subsubsec:standardization}

To preserve discriminative details while maintaining geometric consistency across fonts, character patches were standardized according to their natural proportions:
\begin{itemize}
\item \textbf{Proportional fonts:}
  \begin{itemize}
  \item Wide characters (Q, D, 0, O): $64 \times 32$ pixels
  \item Tall characters (2, Z, S, 5): $32 \times 64$ pixels
  \end{itemize}
\item \textbf{Fixed-width fonts:} $64 \times 64$ pixels
\end{itemize}

All patches were maintained in grayscale with anti-aliasing information preserved to retain subtle shape features essential for discriminating VSCs. This standardization ensures that the binary classifiers receive consistent input dimensions while preserving the aspect ratios critical for character discrimination.

\subsubsection{Dataset statistics and splitting}\label{subsubsec:split}

The extracted character patches were randomly split into training (70\%), validation (20\%), and testing (10\%) sets. This split was performed independently for each VSC pair to ensure balanced representation across all confusable character combinations. For each of the 43 pairs, this yielded approximately 420 training, 120 validation, and 60 testing patches. The split was performed at the image level, so both ROI patches from a given character image were assigned to the same subset.

\subsection{OCR engine selection}\label{subsec:engines}

Several OCR engines were initially considered for this study, including PaddleOCR, Keras-OCR, and commercial APIs. However, some exhibited poor character-level recognition while others were not designed for the single-line text input required by our pipeline. The final selection---Tesseract~4.0, EasyOCR, and TrOCR---represents three widely adopted engines known for high recognition accuracy. Tesseract~4.0 processes text through bidirectional LSTM networks with CTC decoding, achieving 97--98\% character accuracy on clean documents~\cite{smith2007overview,breuel2013high}. EasyOCR combines CRAFT detection with a CRNN architecture (ResNet + BiLSTM + CTC), processing images at 0.4--0.5 seconds per frame with competitive accuracy~\cite{baek2019character,shi2017end,easyocr2020}. TrOCR, a Vision Transformer model with BEiT encoder and RoBERTa decoder, achieves a 2.89\% character error rate on the IAM handwriting benchmark and approximately 96\% word accuracy on SROIE receipts~\cite{li2023trocr}. By selecting engines with such high baseline accuracy, recognition errors in the output strings are predominantly attributable to visually similar character confusion rather than general recognition failures.

\subsection{CVM architecture}\label{subsec:architecture}

\begin{figure}[H]
  \centering
  \includegraphics[width=\linewidth]{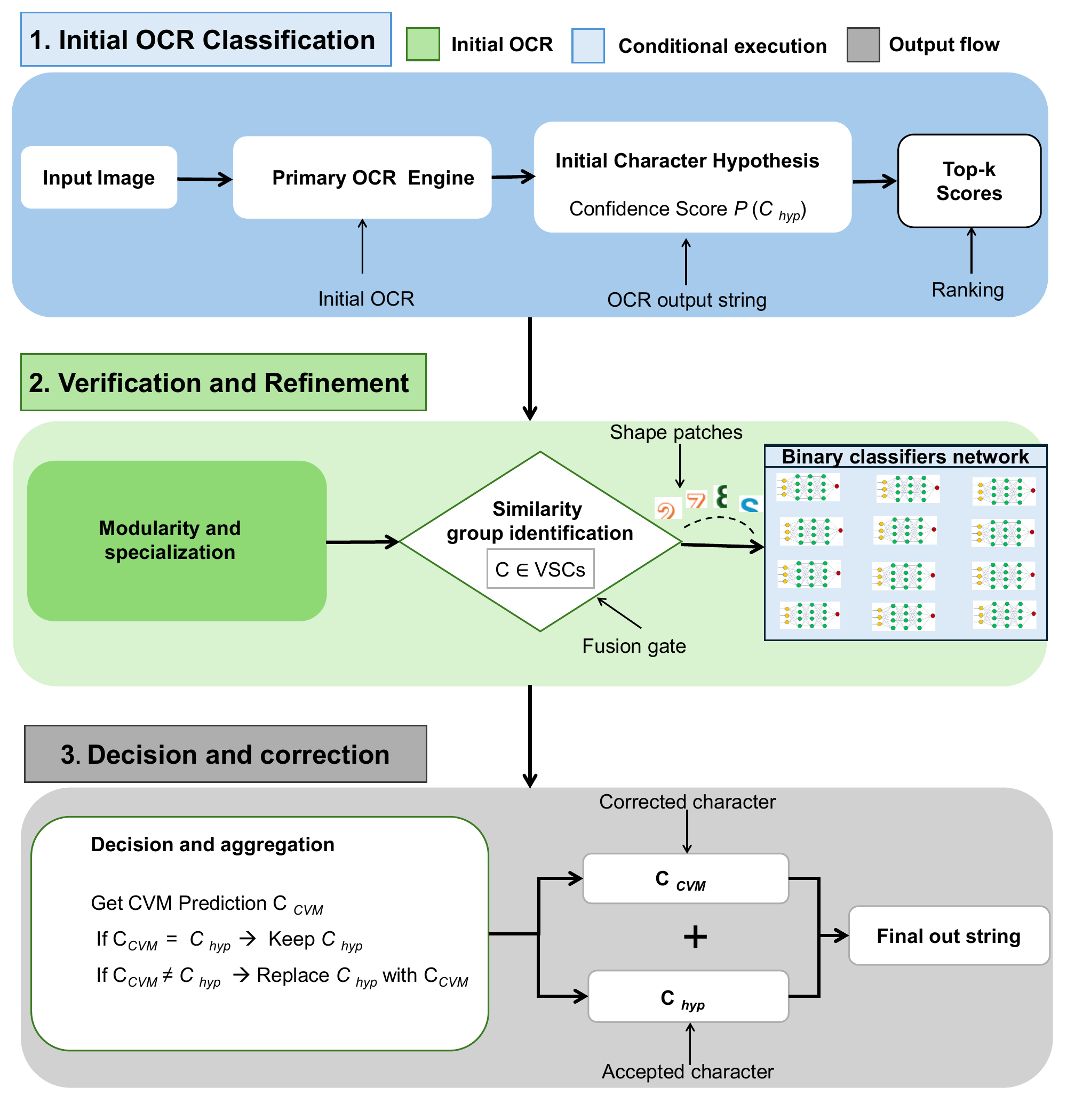}
  \caption{Integrated OCR-CVM workflow showing the complete pipeline.}
  \label{fig:workflow}
\end{figure}

\begin{figure}[t]
  \centering
  \includegraphics[width=0.85\linewidth,height=0.7\textheight,keepaspectratio]{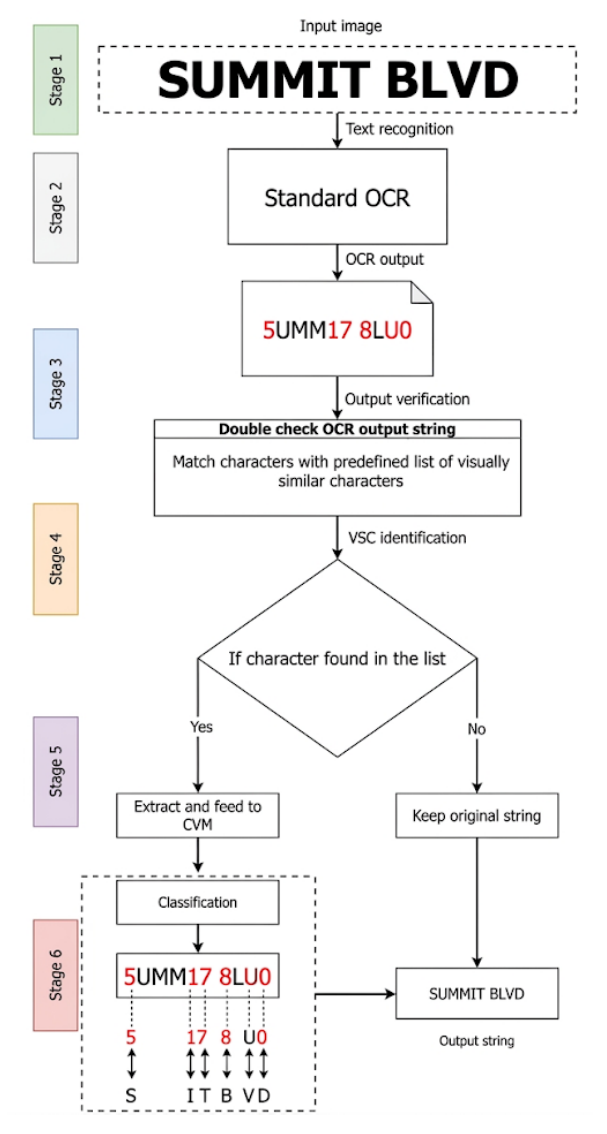}
  \caption{CVM error correction mechanism.}
  \label{fig:mechanism}
\end{figure}

The CVM operates as a post-processing layer integrated with standard OCR engines. The complete workflow, from input image to corrected output, is illustrated in Fig.~\ref{fig:workflow}, while the detailed CVM verification mechanism is shown in Fig.~\ref{fig:mechanism}. The architecture comprises three interconnected modules that work sequentially to identify, verify, and correct visually similar character errors.

\subsubsection{Character localization module}\label{subsubsec:localization}

Given the original input image $I$, the OCR output string $S$, and character bounding boxes $\{B_i\}_{i=1}^{n}$, this module extracts individual character image patches $\{x_i\}_{i=1}^{n}$ corresponding to each recognized character. The module obtains localization information already provided by standard OCR systems, avoiding redundant detection overhead. By extracting patches directly from the original image rather than from preprocessed OCR input, the CVM operates on the same visual information used by the primary recognition system, ensuring consistency in the verification process.

\subsubsection{Confusable character detection}\label{subsubsec:detection}

Before engaging the verification process, the system performs an efficient filtering step. For each character $c_i$ in the output string $S$, we check whether it belongs to any confusable pair in $\mathcal{P}$:
\begin{equation}
c_i \in \bigcup_{j=1}^{m} \{a_j, b_j\}
\label{eq:filter}
\end{equation}
Only characters satisfying this condition are forwarded to the verification stage. This selective processing significantly reduces computational overhead by avoiding unnecessary verification of unambiguous characters that rarely cause confusion. In typical text documents, approximately 30--40\% of characters belong to VSC pairs, meaning the CVM processes only this subset rather than all characters.

\subsubsection{Binary classifier network}\label{subsubsec:binary}

The core of the CVM consists of a collection of specialized binary classifiers $\{C_1, C_2, \ldots, C_m\}$, where each classifier $f_j$ is dedicated to distinguishing between a specific pair of confusable characters $(a_j, b_j)$. When a character $c_i$ is identified as belonging to a VSC pair, the system routes the corresponding character patch $x_i$ to the appropriate binary classifier $C_j$, which outputs both a predicted character $\hat{c}_i$ and an associated confidence score $p_i$.

Each binary classifier is implemented as a Random Forest ensemble for its specific character-pair task. Random Forest was chosen because bagging and random feature selection allow reliable learning from limited image-feature data, unlike deep neural networks~\cite{jalal2022novel}.

By constraining the classification problem to binary decisions between two known confusable characters, these specialized models achieve superior discriminative performance compared to general-purpose multi-class OCR models that must partition the feature space across all possible characters (typically 94 or more classes). The binary decision boundary allows the classifier to focus exclusively on the subtle visual features that distinguish each specific character pair.

\subsection{Training methodology}\label{subsec:training}

\subsubsection{Classifier training}\label{subsubsec:loss}

For each confusable pair $(a_j, b_j)$, we construct a balanced training dataset $\mathcal{D}_j = \{(x_j^{(k)}, y_j^{(k)})\}_{k=1}^{N_j}$, where $x_j^{(k)}$ represents character image patches and $y_j^{(k)} \in \{a_j, b_j\}$ denotes the ground truth label. Each classifier $C_j$ is a Random Forest of $T$ decision trees, each grown on a bootstrap sample of $\mathcal{D}_j$ with node splits selected to minimize Gini impurity over a random subset of features. At inference, the predicted character $\hat{c}_i$ is the majority vote of the trees, and its confidence score is the fraction of trees voting for it:
\begin{equation}
p_i = \frac{1}{T} \sum_{t=1}^{T} \mathbb{1}\left[h_t(x_i) = \hat{c}_i\right]
\label{eq:confidence}
\end{equation}
where $h_t$ denotes the $t$-th tree. This confidence score is compared against the threshold $\tau$ in the correction rule (Eq.~\ref{eq:rule}).

\subsubsection{Training configuration and challenging examples}\label{subsubsec:config}

Critical to effective training is the inclusion of challenging examples that exhibit the visual ambiguities encountered in real-world OCR scenarios. Even under clean imaging conditions and after comprehensive preprocessing, OCR systems misinterpret characters that are intrinsically similar in shape. These ambiguities arise not from noise or degradation, but from genuine visual proximity between certain character forms, an issue that conventional preprocessing is fundamentally unable to resolve.

Therefore, the CVM classifiers are trained on actual confusion patterns observed in OCR outputs in multiple fonts. By learning from these empirically grounded cases of character-level ambiguity, the model acquires the discriminative ability to separate confusable shapes and correct persistent errors in standard OCR pipelines. Each Random Forest classifier was configured with hyperparameters optimized during validation to maximize performance on the specific character pair it addresses. All training was performed in Python on a system equipped with a 12th Gen Intel\textregistered{} Core\texttrademark{} i5-12400F processor (2.50\,GHz) and 64\,GB of RAM, running a 64-bit operating system. The training data split (70\% training, 20\% validation, 10\% testing) ensures sufficient samples for model development while preserving held-out data for unbiased performance evaluation.

\subsubsection{Classifier independence and modularity}\label{subsubsec:modularity}

\begin{figure}[t]
  \centering
  \includegraphics[width=0.8\linewidth]{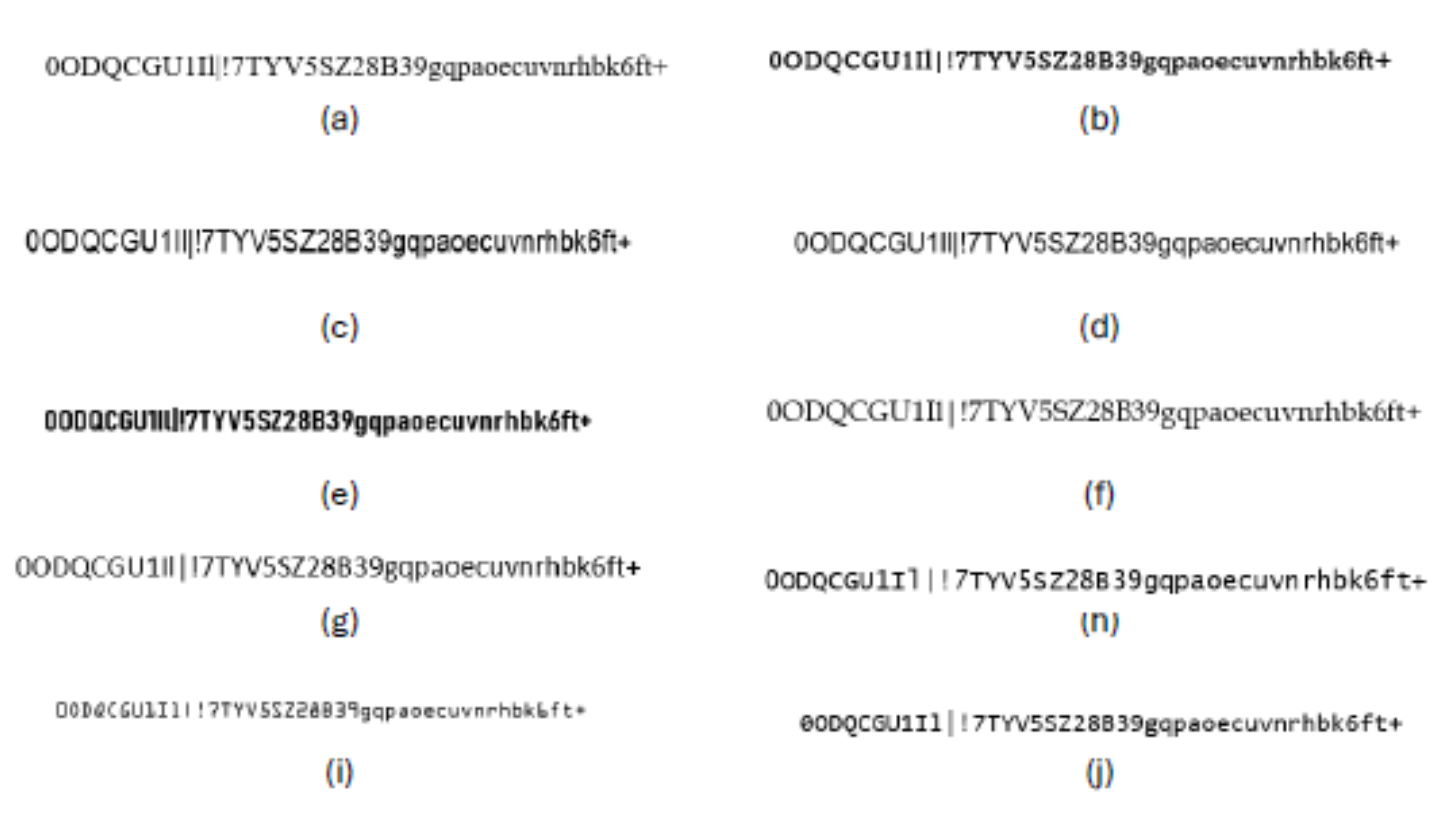}
  \caption{Sample unseen test images containing random strings in 10 different fonts.}
  \label{fig:fonts}
\end{figure}

Training proceeds independently for each binary classifier for parallel optimization and modular deployment. This independence has several advantages: (1) classifiers can be trained in parallel, (2) poor performance of one classifier does not affect others, (3) new confusable pairs can be added by training additional classifiers without retraining existing ones, and (4) individual classifiers can be updated or fine-tuned based on application-specific error patterns. This modular architecture makes the system easily extensible to accommodate new use cases, fonts, or languages. Unseen images containing random strings in 10 different fonts were used to evaluate the performance of both the baseline and CVM-enhanced OCR engines (Fig.~\ref{fig:fonts}).

\subsection{Inference pipeline}\label{subsec:inference}

\subsubsection{Verification algorithm}\label{subsubsec:algorithm}

At inference time, CVM operates as a post-processing module receiving three inputs: (1) the original image $I$, (2) the OCR output string $S$, and (3) character bounding boxes $\{B_i\}_{i=1}^{n}$. The verification proceeds according to Algorithm~\ref{alg:cvm}:
\begin{enumerate}
\item Extract character image patches $\{x_i\}_{i=1}^{n}$ from the original image using OCR-provided bounding boxes.
\item For each character $c_i$ in the OCR output string $S$:
  \begin{enumerate}
  \item Check if $c_i$ belongs to any VSC pair in $\mathcal{P}$.
  \item If true, identify the corresponding pair $(a_j, b_j)$ and its associated classifier $C_j$.
  \item Feed the character patch $x_i$ to classifier $C_j$ to obtain verified character $\hat{c}_i$ and confidence score $p_i$.
  \item Apply the correction rule given as:
  \begin{equation}
  c_i^{\text{final}} =
  \begin{cases}
  \hat{c}_i & \text{if } \hat{c}_i \neq c_i \text{ and } p_i \geq \tau \\
  c_i & \text{otherwise}
  \end{cases}
  \label{eq:rule}
  \end{equation}
  where $\tau$ is the confidence threshold.
  \end{enumerate}
\item Reconstruct the corrected output string $\hat{S}$ using the finalized characters.
\end{enumerate}

The corrected string $\hat{S}$ preserves all characters where either (1) the character does not belong to any VSC pair, (2) the CVM agrees with the original OCR prediction, or (3) the CVM disagrees but has insufficient confidence ($p_i < \tau$) to override the original prediction. Character accuracy is computed as the number of correctly recognized characters divided by the number of characters in the ground-truth string, after aligning the OCR output to the ground truth.

\begin{algorithm}[H]
\caption{Character Verification Model (CVM)}\label{alg:cvm}
\begin{algorithmic}[1]
\Require Image $I$, OCR output $S = c_1 \ldots c_n$, bounding boxes $\{B_i\}_{i=1}^{n}$, VSC pairs $\mathcal{P}$, classifiers $\{C_j\}_{j=1}^{m}$
\Ensure Corrected string $\hat{S}$
\State Extract character patches $\{x_i\}_{i=1}^{n}$ from $I$ using $\{B_i\}_{i=1}^{n}$
\For{each character $c_i$ in $S$}
  \If{$c_i \in \{a_j, b_j\}$ for some $(a_j, b_j) \in \mathcal{P}$}
    \State Identify corresponding classifier $C_j$
    \State $(\hat{c}_i, p_i) \gets f_j(x_i)$
    \If{$(\hat{c}_i \neq c_i)$ \textbf{and} $(p_i \geq \tau)$}
      \State $c_i^{\text{final}} \gets \hat{c}_i$
    \Else
      \State $c_i^{\text{final}} \gets c_i$
    \EndIf
  \Else
    \State $c_i^{\text{final}} \gets c_i$
  \EndIf
\EndFor
\State Construct corrected string $\hat{S}$ from all $\{c_i^{\text{final}}\}_{i=1}^{n}$
\State \Return $\hat{S}$
\end{algorithmic}
\end{algorithm}

\subsection{System integration and compatibility}\label{subsec:integration}

A key design principle of CVM is its modularity and compatibility with existing OCR systems. The module functions as a standalone post-processing layer requiring only three standard outputs from any OCR engine: the original image, the recognized text string, and character-level bounding boxes. These outputs are provided by most modern OCR systems (including Tesseract, EasyOCR, TrOCR, and commercial APIs), making CVM readily integrable without any modifications to the underlying OCR architecture.

There are several practical advantages of the modular design. First, the same CVM implementation can be deployed with different OCR engines for consistent verification capabilities regardless of the primary recognition system used. Second, the binary classifiers can be trained and updated independently, enabling incremental improvements without system-wide retraining. Third, new confusable character pairs can be added simply by training and incorporating additional binary classifiers, making the system easily extensible to accommodate new fonts, languages, or domain-specific character sets. This architecture ensures seamless deployment as an enhancement layer for existing OCR pipelines in production environments.

\section{Results and discussion}\label{sec:results}

We evaluated CVM through three OCR engines (Tesseract, EasyOCR, TrOCR) and ten fonts with varying typographic characteristics. The evaluation proceeded in three phases: baseline performance assessment to identify error patterns, classification of errors by correctability, and measurement of CVM performance improvements.

\subsection{Baseline performance and error characteristics}\label{subsec:baseline}

\begin{figure}[H]
\centering
\begin{subfigure}[b]{0.48\textwidth}
  \centering
  \includegraphics[width=\textwidth]{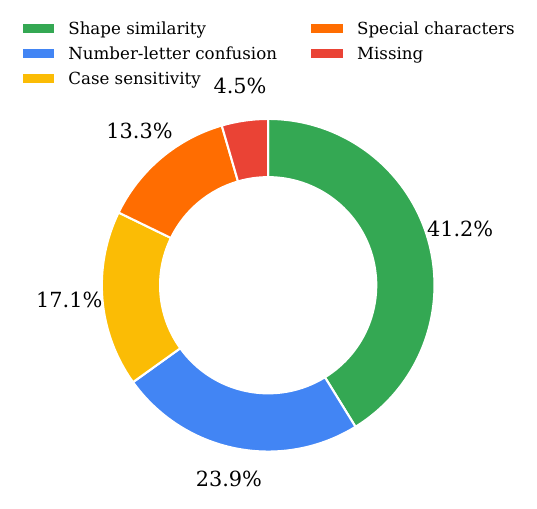}
  \caption{Character recognition errors of baseline OCR engines}
  \label{fig:error_categories}
\end{subfigure}\hfill
\begin{subfigure}[b]{0.48\textwidth}
  \centering
  \includegraphics[width=\textwidth]{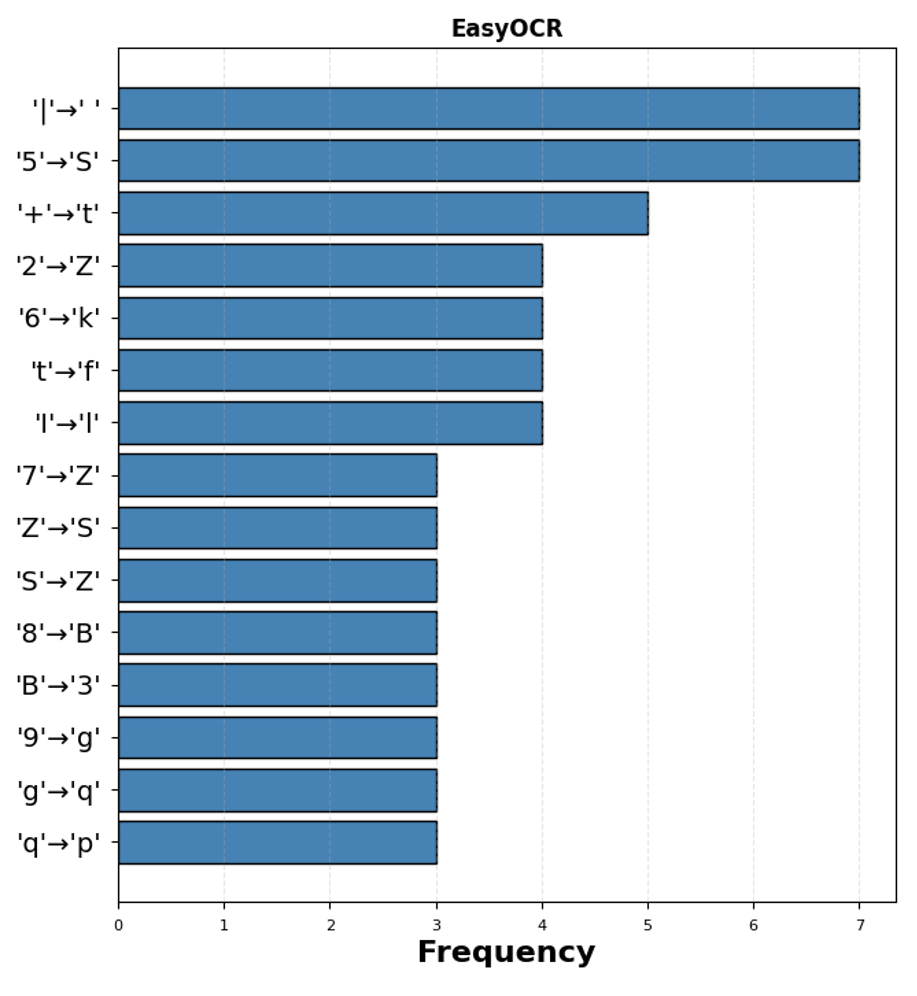}
  \caption{Frequent misclassification patterns in EasyOCR}
  \label{fig:easyocr_errors}
\end{subfigure}

\vspace{0.5em}
\begin{subfigure}[b]{0.48\textwidth}
  \centering
  \includegraphics[width=\textwidth]{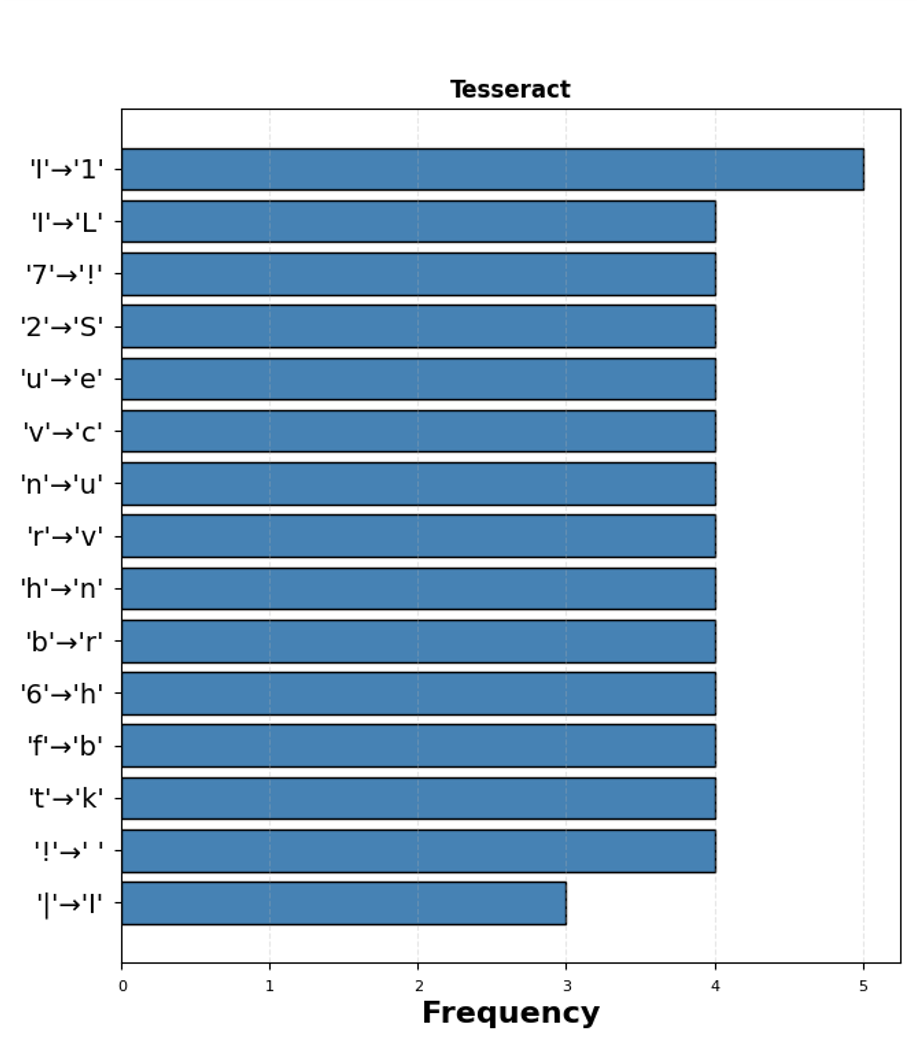}
  \caption{Frequent misclassification patterns in Tesseract}
  \label{fig:tesseract_errors}
\end{subfigure}\hfill
\begin{subfigure}[b]{0.48\textwidth}
  \centering
  \includegraphics[width=\textwidth]{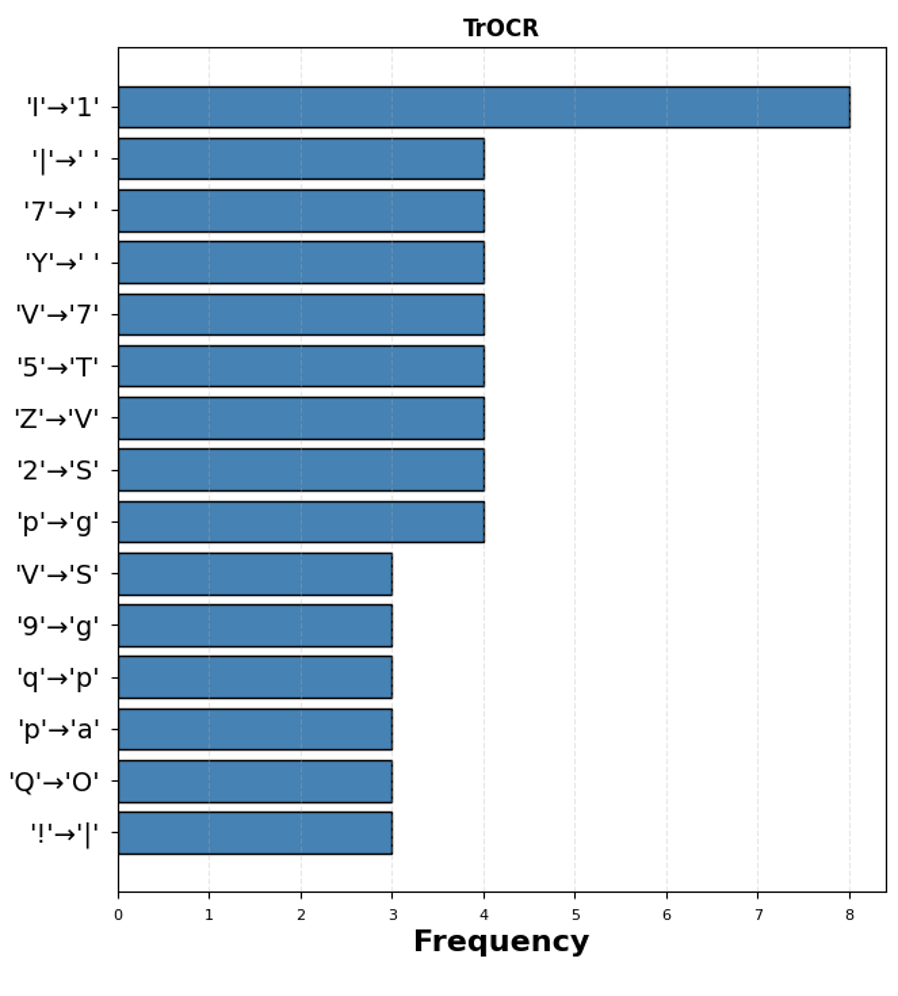}
  \caption{Frequent misclassification patterns in TrOCR}
  \label{fig:trocr_errors}
\end{subfigure}

\caption{Evaluation of recognition issues in OCR engines.}
\label{fig:errors}
\end{figure}

Initial analysis of OCR outputs (Fig.~\ref{fig:errors}a) revealed that visual similarity between characters accounts for 41.2\% of all recognition failures, representing the single largest error category. Number--letter confusion contributes an additional 23.9\%, while special character misrecognition accounts for 17.1\%, missing characters 13.3\%, and case sensitivity issues 4.5\% of errors. The substantial proportion attributable to visual similarity and number--letter confusion (collectively 65.1\% of failures) validates the targeted character-level approach, suggesting that focused verification of confusable pairs may be more effective than broader contextual correction methods. Examination of engine-specific confusion patterns (Figs.~\ref{fig:errors}b--d) revealed several consistent trends. The character `I' (uppercase i) was frequently substituted as `1', `l', or `|', representing the most common single misrecognition pattern. Rounded characters showed predictable confusion patterns, with lowercase `o' commonly misrecognized as `e', `a', or `c', while uppercase `O' appeared as `0', `o', `D', or `@' depending on font characteristics. Among alphanumeric pairs, the most frequent confusions involved `S'/`5', `Z'/`2', `B'/`8', and `g'/`q', pairs that exhibited strong visual similarity across multiple fonts. Characters with limited geometric features produced high cross-category confusion, with underscore replaced by space, `+' by `t' or `f', and `-' by space.

\begin{figure}[H]
  \centering
  \includegraphics[width=0.85\linewidth]{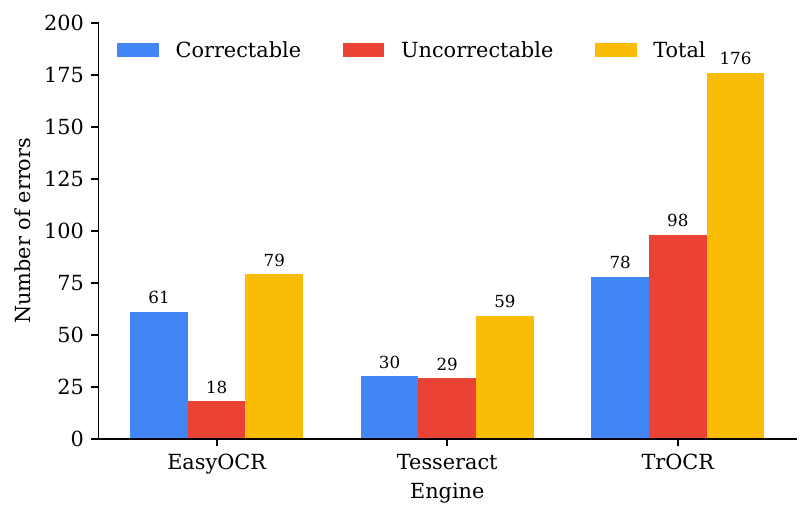}
  \caption{Distribution of correctable and uncorrectable errors across OCR engines.}
  \label{fig:correctable}
\end{figure}

Before deploying post-processing, we examined what proportion of OCR errors falls within the system correction scope. CVM addresses substitution errors involving predefined VSC pairs but cannot correct insertion errors (spurious characters), deletion errors (missing characters), multi-character segmentation errors (e.g., `rn' merged as `m'), or substitutions outside the VSC pair set. Analysis revealed that 53.8\% of all errors (169 of 314) were correctable, while 46.2\% fell outside CVM scope (Fig.~\ref{fig:correctable}). However, this proportion varied significantly by engine: EasyOCR showed 77.2\% correctable errors (61 of 79), Tesseract 50.8\% (30 of 59), and TrOCR 44.3\% (78 of 176). The higher proportion for EasyOCR suggests its misrecognitions are predominantly character-level substitutions involving visually similar pairs, while TrOCR exhibits a more diverse error profile including segmentation issues and non-VSC substitutions requiring different correction strategies. These findings establish the theoretical upper bound for CVM impact: if it corrected 100\% of VSC errors with no false corrections, overall error rates could decrease by 53.8\% (relative reduction).

\subsection{CVM performance improvements}\label{subsec:improvements}

\begin{figure}[H]
  \centering
  \includegraphics[width=0.85\linewidth]{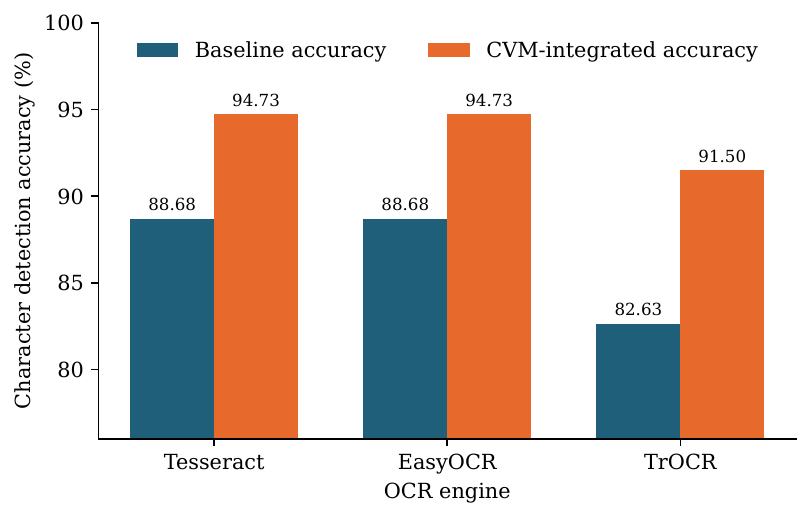}
  \caption{Overall OCR accuracy before and after CVM integration.}
  \label{fig:accuracy}
\end{figure}

Integrating CVM as a post-processing layer consistently improved accuracy for all tested engines (Fig.~\ref{fig:accuracy}), with baseline accuracy ranging from 73.68\% to 92.11\% across engines and fonts (Table~\ref{tab:fonts}). Averaged across the ten test fonts, Tesseract improved from a baseline of 88.68\% to 94.73\% (+6.05 percentage points), EasyOCR showed identical improvement from 88.68\% to 94.73\%, while TrOCR demonstrated the largest gain from 82.63\% to 91.50\% (+8.87 percentage points). The similar performance of Tesseract and EasyOCR is attributable to both engines employing comparable character-level recognition architectures and exhibiting similar confusion patterns with VSC pairs. TrOCR, a transformer-based model, showed substantially larger absolute improvement, suggesting that attention-based architectures may struggle more with fine-grained visual discrimination of confusable characters compared to traditional CNN-based approaches. This pattern likely reflects that transformer models rely more heavily on global context than local feature discrimination, making them more susceptible to character-level visual ambiguity. Relative to baseline error rates, these gains represent error reductions of 53.4\% for Tesseract and EasyOCR (from 11.32\% error to 5.27\%) and 51.1\% for TrOCR (from 17.37\% error to 8.50\%). Tesseract and EasyOCR produced identical results for every font because both engines made the same misrecognitions at the same character positions on each test string; consequently, CVM received identical inputs from both engines and applied identical corrections. The improvement was consistent across all ten fonts for every engine; an exact two-sided Wilcoxon signed-rank test on the per-font accuracies before and after CVM gives $p = 0.002$ for each of Tesseract, EasyOCR, and TrOCR ($n = 10$ fonts, all gains positive).

Detailed examination of the correction mechanism, aggregated across all evaluation runs (Table~\ref{tab:stats}), revealed that of 876 VSC errors in baseline OCR outputs, CVM successfully corrected 765 (87.3\%) and missed 111 (12.7\%). CVM also introduced 18 false corrections, i.e., 2.3\% of the 783 corrections it applied, yielding a net error reduction of 747. Note that Fig.~\ref{fig:correctable} reports errors from a single evaluation pass, whereas Table~\ref{tab:stats} aggregates over all runs. This high correction rate combined with a low false-correction rate demonstrates the effectiveness and precision of verification-based correction. Several factors contribute to this performance. The constrained decision space enables superior discrimination, as each binary classifier distinguishes between only two known confusable characters rather than performing full multi-class recognition. By operating on character patches extracted from the original image, CVM exploits the same visual information available to the primary OCR engine but with specialized training focused on specific confusable pairs. The 70\% confidence threshold ensures that only high-certainty corrections are applied, maintaining a low false-correction rate while allowing most genuine errors to be fixed. Finally, each binary classifier trains exclusively on challenging examples of its specific character pair, including actual confusion cases observed in OCR outputs across multiple fonts, enabling the model to learn discriminative features that general-purpose classifiers may overlook.

\begin{table}[t]
\caption{Overall CVM correction statistics, aggregated across all evaluation runs.}\label{tab:stats}
\centering
\begin{tabular}{@{}lr@{}}
\toprule
Metric & Value \\
\midrule
Total VSC errors (baseline) & 876 \\
Successfully corrected by CVM & 765 \\
Missed VSC errors & 111 \\
\textbf{VSC correction rate} & \textbf{87.3\%} \\
False corrections introduced & 18 \\
Total corrections applied & 783 \\
\textbf{False correction rate} & \textbf{2.3\%} \\
Net error reduction & 747 \\
\botrule
\end{tabular}
\end{table}

Analysis of performance across fonts (Table~\ref{tab:fonts}) revealed patterns related to typographic characteristics. For Tesseract and EasyOCR, sans-serif fonts with geometric properties (Arial Nova, Bahnschrift SemiBold) exhibited lower baseline accuracy (84.21--86.84\%) and larger improvements from CVM (6.46--10.25 percentage points), while TrOCR showed the same trend more strongly (baseline 76.32--81.58\%, gains of 9.66--14.81 percentage points). Geometric uniformity in character shapes increases visual similarity, which elevates baseline confusion rates while leaving discriminative features that the binary classifiers can exploit. Serif fonts (Times New Roman, Book Antiqua) yielded higher Tesseract and EasyOCR baselines (89.47--92.11\%) and smaller improvements (4.12--4.34 percentage points), consistent with serif ornamentation providing additional visual cues that reduce initial confusion. TrOCR did not follow this pattern: its lowest baseline (73.68\%) and second-largest gain (+13.56 percentage points) occurred on Times New Roman, indicating that font effects are engine-dependent. Monospace fonts (Consolas, Lucida Console) gave intermediate results, with Tesseract and EasyOCR baselines of 89.47\% and improvements of 4.36--5.10 percentage points (TrOCR: 86.84\% baseline, 5.99--6.66 percentage points). The OCR-optimized font OCR Extended showed counterintuitive results, with low baselines (81.58\% for Tesseract and EasyOCR, 76.32\% for TrOCR) but among the largest improvements (11.90 and 14.55 percentage points, respectively). This likely reflects that OCR Extended was designed for older recognition systems; modern OCR engines may not be specifically tuned for this font, yet its exaggerated character differentiation provides strong discrimination features for the binary classifiers.

\begin{table}[!htbp]
\caption{Character accuracy (\%) before and after CVM integration on the ten test fonts. Tesseract and EasyOCR produced identical results on every font and are reported together.}\label{tab:fonts}
\centering
\small
\begin{tabular}{@{}lrrrrrr@{}}
\toprule
 & \multicolumn{3}{c}{Tesseract / EasyOCR} & \multicolumn{3}{c}{TrOCR} \\
\cmidrule(lr){2-4}\cmidrule(l){5-7}
Font & Before & After & Gain & Before & After & Gain \\
\midrule
Times New Roman      & 92.11 & 96.45 & +4.34  & 73.68 & 87.24 & +13.56 \\
Consolas             & 89.47 & 94.57 & +5.10  & 86.84 & 93.50 & +6.66 \\
Amasis MT Pro Med    & 92.11 & 96.06 & +3.95  & 86.84 & 93.42 & +6.58 \\
Arial Narrow         & 89.47 & 95.14 & +5.67  & 86.84 & 93.81 & +6.97 \\
Arial Nova           & 86.84 & 93.30 & +6.46  & 81.58 & 91.24 & +9.66 \\
Bahnschrift SemiBold & 84.21 & 94.46 & +10.25 & 76.32 & 91.13 & +14.81 \\
Book Antiqua         & 89.47 & 93.59 & +4.12  & 81.58 & 88.62 & +7.04 \\
Calibri              & 92.11 & 96.42 & +4.31  & 84.21 & 92.37 & +8.16 \\
Lucida Console       & 89.47 & 93.83 & +4.36  & 86.84 & 92.83 & +5.99 \\
OCR Extended         & 81.58 & 93.48 & +11.90 & 76.32 & 90.87 & +14.55 \\
\midrule
Mean                 & 88.68 & 94.73 & +6.05  & 82.11 & 91.50 & +9.39 \\
\botrule
\end{tabular}
\end{table}

\subsection{System limitations}\label{subsec:limitations}

While CVM demonstrates substantial improvements in correcting VSC errors, several limitations define its operational boundaries. The single-character restriction means CVM cannot address multi-character segmentation errors where character boundaries are incorrectly detected. Common examples include `rn' misrecognized as `m', `cl' as `d', `13' as `B', and `ii' as `u'. Since CVM receives OCR output as input, it inherits segmentation boundaries and cannot detect or correct such errors. By design, CVM addresses only substitution errors and cannot detect or correct insertion, deletion, or transposition errors. Comprehensive post-processing would require CVM to operate as one component in a multi-stage correction pipeline addressing different error types through complementary mechanisms. Certain character pairs exhibit near-identical or truly identical shapes in specific fonts, making visual discrimination impossible even for specialized binary classifiers. In sans-serif fonts like Arial and Calibri, uppercase `I', lowercase `l', and digit `1' are visually indistinguishable. For these truly homoglyphic pairs, even human readers need contextual information for disambiguation, placing them fundamentally beyond the reach of appearance-based verification. Performance variation across font families suggests that domain-specific deployment may benefit from classifier fine-tuning. The evaluation also relies on synthetically rendered strings with clean backgrounds; performance on scanned, photographed, or degraded images, where character segmentation is less reliable, remains to be established. Finally, several evaluation fonts belong to font families that also appear in the 94-class training dataset, so results on entirely unseen typefaces may be lower. The 70\% confidence threshold employed balanced correction accuracy against false-correction risk, but different applications may require adjustment for specific risk profiles.

\section{Conclusion}\label{sec:conclusion}

Distinguishing visually similar characters (VSCs) has long been recognized as a fundamental challenge in OCR, yet prior work has addressed it primarily through improved preprocessing or more powerful recognition models. This study introduces an alternative approach: accepting that even state-of-the-art models will confuse VSCs, and systematically verifying such characters through specialized binary classifiers as a post-processing step.

The empirical results demonstrate that this verification-based paradigm delivers substantial accuracy improvements for diverse OCR engines and fonts, particularly for non-lexical content where dictionary-based post-processing methods fail entirely. By operating at the character level with vocabulary-independent visual discrimination, CVM fills a critical gap in OCR correction capabilities and extends reliable text recognition to applications previously constrained by persistent VSC-related errors.

As text digitization continues expanding into safety-critical and business-critical domains, the ability to systematically verify and correct ambiguous characters without requiring linguistic context or domain-specific vocabularies represents an important step toward more reliable and broadly applicable OCR systems. The verification template for targeted error correction complements existing approaches and addresses currently unaddressed error categories despite advances in OCR. Future work will extend these experiments to real-world mixed-font and non-lexical datasets as they become available.

\section*{CRediT authorship contribution statement}
\textbf{Nuzhat Khan:} Conceptualization, Methodology, Software, Investigation, Formal analysis, Visualization, Writing -- original draft.
\textbf{Ab Al-Hadi Ab Rahman:} Conceptualization, Supervision, Funding acquisition, Project administration, Writing -- review \& editing.
\textbf{Mohamad Anuar Kamaruddin:} Supervision, Resources, Writing -- review \& editing.
\textbf{Muhammad Paend Bakht:} Validation, Formal analysis, Writing -- review \& editing.
\textbf{Sana Fatima:} Coding, Writing.
\textbf{Farhad Khodadadi:} Writing -- review \& editing..
\textbf{Ibrahim Yousef Alshareef:} Writing -- review \& editing.

\section*{Declaration of competing interest}
The authors declare that they have no known competing financial interests or personal relationships that could have appeared to influence the work reported in this paper.

\section*{Data availability}
The character dataset used for training is publicly available at \url{https://github.com/VeCAD-P04-UTM/94-Class-ASCII-Image-OCR-Data}. Test images and OCR outputs will be made available on request.

\section*{Declaration of generative AI and AI-assisted technologies in the writing process}
During the preparation of this work the authors used Claude (Anthropic) and Gemini (Google) in order to reconstruct the LaTeX source, language editing, and proofreading. After using these tools, the authors reviewed and edited the content as needed and take full responsibility for the content of the publication.

\section*{Acknowledgements}
This work was supported in part by the Universiti Teknologi Malaysia with the Flagship CoE/RG Research under Grant Q.J130000.5023.10G05 and Grant Q.J130000.5023.10G06, and in part by the Professional Development under Grant R.J130000.7113.07E51.
\clearpage
\bibliographystyle{elsarticle-num}
\bibliography{references}

\end{document}